\documentclass{article}

\usepackage[preprint]{neurips_2026}

\usepackage[utf8]{inputenc} 
\usepackage[T1]{fontenc}    
\usepackage{url}            
\usepackage{nicefrac}       

\usepackage{amsmath,amssymb,amsfonts}

\usepackage{newtxtext,newtxmath}
\usepackage{textcomp}
\usepackage{microtype}
\usepackage{soul}
\usepackage{comment}
\usepackage{lipsum}

\usepackage{xcolor}
\usepackage{colortbl}

\usepackage{array}
\usepackage{booktabs}
\usepackage{makecell}
\usepackage{multirow}
\usepackage{tabularray}
\usepackage{arydshln}
\usepackage{tablefootnote}

\usepackage{graphicx}
\usepackage[export]{adjustbox}
\usepackage{wrapfig}
\usepackage{float}
\usepackage{subcaption}
\usepackage{svg}

\usepackage{enumitem}
\usepackage{outlines}

\usepackage{pifont}
\usepackage{fontawesome5}
\usepackage{listings}
\usepackage{tcolorbox}
\usepackage{mdframed}
\usepackage[colorinlistoftodos,textsize=small]{todonotes}
\usepackage{xurl}
\usepackage{tikz}
\usetikzlibrary{shapes,arrows,positioning}
\usepackage{forest}

\usepackage{fancyhdr}

\usepackage{hyperref}
\usepackage[nameinlink,capitalise,noabbrev]{cleveref}

\crefformat{section}{\S#2#1#3}
\crefformat{subsection}{\S#2#1#3}
\crefformat{subsubsection}{\S#2#1#3}

\usepackage{hyperref}       

\title{How Far Can Disaggregation Go? A Design-Space Exploration of Attention–FFN Disaggregation for Efficient MoE LLM Serving}

\author{%
  \normalfont
  Hanjiang Wu$^{1}$ \quad
  Abhimanyu Rajeshkumar Bambhaniya$^{1,5}$ \quad
  Sarbartha Banerjee$^{1}$ \\
  Tuhin Khare$^{1}$ \quad
  Sudarshan Srinivasan$^{2}$ \quad
  Suvinay Subramanian$^{3}$ \\
  Souvik Kundu$^{2}$ \quad
  Madhu Kumar$^{2}$ \quad
  Midhilesh Elavazhagan$^{2}$ \\
  William Won$^{1}$ \quad
  Amir Yazdanbakhsh$^{4}$ \quad
  Tushar Krishna$^{1,5}$ \\[4pt]
  \normalfont
  $^{1}$Georgia Institute of Technology \quad
  $^{2}$Intel \quad
  $^{3}$Google \quad
  $^{4}$Google DeepMind \quad
  $^{5}$Infravana \\
}

\begin{document}

\maketitle

\begin{abstract}

Modern large language model (LLM) inference has progressively disaggregated to keep pace with growing model sizes and tight \texttt{TTFT} and \texttt{TPOT} service-level objectives: from chunked-prefill aggregation, to prefill--decode (P/D) disaggregation, and most recently to operator-level Attention--FFN Disaggregation (AFD). This trend is especially important for mixture-of-experts (MoE) models, where memory-bound attention, compute-intensive expert FFNs, and MoE dispatch/combine communication create distinct resource demands across the serving pipeline. AFD further exposes this heterogeneity by placing attention and MoE-FFN execution on separate GPU groups. Each level of disaggregation deepens the scheduling design space across workload characteristics, resource allocation, and interconnect topology, leaving open the central question: \textit{When does each level of disaggregation actually pay off?} We systematically characterize this trade-off for MoE inference across realistic workload use cases defined by input/output sequence lengths, prefix-KV reuse, and per-user latency constraints. Using chunked-prefill and P/D disaggregation as strong baselines, we study the benefits and limits of AFD at scale through a framework that fuses rich on-device kernel measurements with high-fidelity network simulation. Our findings deliver a practical map of when and where deeper disaggregation pays off for MoE serving at scale. Under strict \texttt{TTFT/TPOT} SLOs, AFD sustains around 4k tokens/s of system throughput on DeepSeek-V3.2 across chat, coding, and agentic-coding workloads, regimes in which non-AFD deployments are infeasible. Our design and analysis further distill concrete takeaways for jointly optimizing system throughput and user interactivity, including how to partition attention and FFN across GPUs as a function of workload and model architecture, providing design principles for current rack- and cluster-scale deployments as well as future disaggregated AI infrastructure.

\end{abstract}

\section{Introduction}\label{sec:intro}

The rapid scaling of agentic large language models (LLMs) has enabled AI systems to perform increasingly complex tasks, including multi‑turn reasoning, code generation, and autonomous decision‑making. As these models grow to hundreds of billions of parameters and operate on long input contexts, their deployment places unprecedented pressure on inference infrastructure, particularly due to the expanding KV‑cache footprint and the need to scale across multiple compute nodes. At the same time, emerging agentic workloads and model architectures exhibit increasing compute characteristic heterogeneity, exposing limitations in existing LLM serving paradigms that struggle to simultaneously achieve high performance, efficiency, and scalability.

A central challenge stems from the heterogeneous execution characteristics of different components within modern LLM architectures, which impose conflicting demands on compute, memory bandwidth, and communication resources. Prior systems mitigate these effects through batching and scheduling techniques such as chunked prefill~\cite{agrawal2023sarathi} and continuous batching~\cite{yu2022orca}, pipelining, or coarse‑grained prefill–decode (P/D) disaggregation~\cite{zhong2024distserve, patel2023splitwise}. While effective at reducing phase‑level interference, these approaches implicitly treat the model as a monolithic execution unit, obscuring fine‑grained trade‑offs that become dominant at scale.

\begin{figure}[t]
    \includegraphics[width=1\linewidth, trim=0.85cm 0 0 0, clip]
{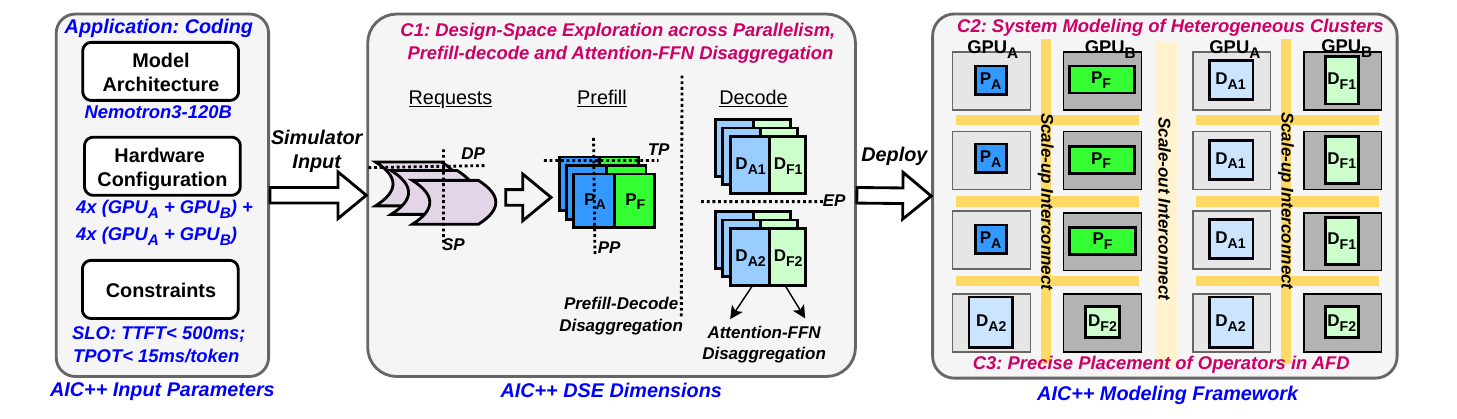}
    \vspace{-1em}
\caption{\textbf{\texttt{AIC++} Framework Overview.} \texttt{AIC++} takes model architecture, hardware configuration, and workload constraints as inputs and performs design-space exploration to identify optimal scheduling across token-level parallelism—data (DP), sequence (SP), tensor (TP), pipeline (PP), and expert parallelism (EP)—as well as phase-level prefill/decode (P/D) and operator-level attention--FFN disaggregation (AFD) \textbf{[C1]}. The framework leverages AIConfigurator to model heterogeneous GPU clusters (\(GPU_A, GPU_B\)) interconnected via scale-up (NVLink) and scale-out (InfiniBand) fabrics simulated with AstraSim \textbf{[C2]}. Based on this analysis, attention (\(P_A, D_{A1}, D_{A2}\)) and FFN (\(P_F, D_{F1}, D_{F2}\)) operators are placed on specific GPUs to maximize overall system throughput \textbf{[C3]}.}
    \vspace{-2em}
    \label{fig:dse}
\end{figure}

Recent studies such as MegaScale-Infer~\cite{zhu2025megascaleinfer} show that coarse-grained inference abstractions break down for large heterogeneous models, especially MoE architectures, where substantial compute heterogeneity exists within each Transformer block. Attention variants such as MHA~\cite{ashish2017attention}, GQA~\cite{hudson2019gqa}, and MLA~\cite{deepseekai2024deepseekv2} are largely memory-bound due to KV-cache access and data movement, whereas FFNs are compute-bound and dominated by dense GEMMs. Although MegaScale-Infer highlights inefficiencies from attention--FFN heterogeneity, it remains unclear how Attention--FFN Disaggregation (AFD) composes with existing parallelism strategies, prefill--decode (P/D) disaggregation, and different attention architectures.

Beyond architectural heterogeneity, AFD effectiveness also depends on workload characteristics such as input/output sequence length (ISL/OSL), prefix length, and system load (\texttt{tokens/s/user}). These factors directly affect scheduling decisions and determine when disaggregation is beneficial. Understanding these trade-offs is critical for current cluster-scale LLM serving and for future disaggregated inference platforms, including NVIDIA Groq 3 LPX~\cite{nvidia_lpx} and Intel/SambaNova-style systems~\cite{sambanova}.

In this paper, we present a systematic study of AFD for LLM inference across diverse application domains. We analyze efficient deployment across three dimensions: distributed parallelism, including tensor, data, pipeline, sequence, and expert parallelism; phase-level P/D disaggregation; and operator-level attention--FFN disaggregation, as shown in~\cref{fig:dse}\textbf{(C1)}. We formulate scheduling selection as a design-space exploration (DSE) problem that captures operator-level compute heterogeneity and inter-node communication costs.

To support this study, we develop AIConfigurator++ (\texttt{AIC++}), a co-design framework that combines operator-level compute modeling from NVIDIA AIConfigurator~\cite{xu2026aic} with distributed communication modeling using AstraSim~\cite{astrasim}. Grounded in a customized vLLM-based AFD prototype~\cite{kwon2023vllm}, \texttt{AIC++} integrates kernel measurements with system-level simulation to evaluate compute--communication trade-offs across scheduling strategies, as illustrated in~\cref{fig:dse}\textbf{(C2)}.

We evaluate chatbot, coding, and agentic workloads using \texttt{DeepSeek-V3.2}, \texttt{GPT-OSS-120B}, \texttt{Nemotron3-120B}, and \texttt{Qwen3-235B}, covering MLA, GQA, sparse attention, and Mamba-based architectures. By varying ISL, OSL, prefix length, and system load, our framework identifies when AFD is beneficial, how micro-batching and operator placement improve compute--communication overlap, and which scheduling strategy maximizes throughput and interactivity under SLO constraints. Importantly, our analysis translates workload and model characteristics into concrete attention-to-FFN GPU ratios, providing actionable guidance for cluster-scale deployment.

\begin{figure}[t]
    \centering
    \includegraphics[width=\linewidth]{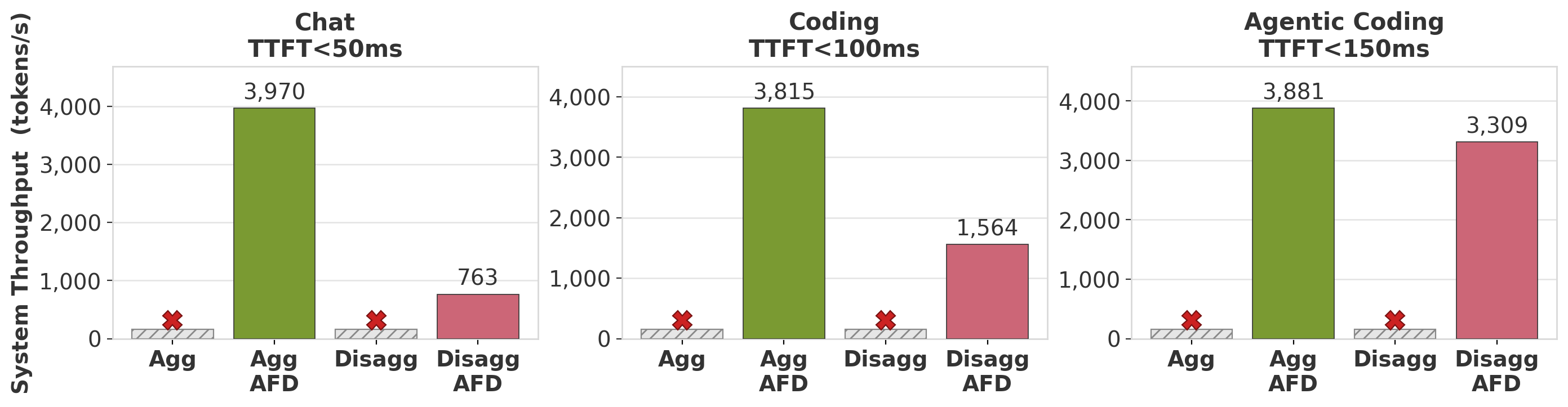}
    \vspace{-1.5em}
    \caption{System throughput at the best feasible deployment on \texttt{DeepSeek-V3.2} (128× B200 with trtllm backend) under strict SLOs (\texttt{TTFT<50/100/150 ms} for Chat/Coding/Agentic Coding; \texttt{TPOT caps 15 ms}). Red cross marks infeasibility — the auto-config search finds no deployment that satisfies the SLO. Only AFD-attention variants (\texttt{Agg+AFD, P/D Disagg+AFD}) unlock feasible configurations, sustaining up to  $\sim$ \texttt{4K} tokens/s for the overall system throughput.}
    \vspace{-1.5em}
    \label{fig:deepseekv32_strict_128gpu}
\end{figure}

More broadly, our findings suggest that as LLM workloads become increasingly heterogeneous, system optimization must move beyond coarse-grained placement toward operator-level disaggregation. Modeling-driven frameworks such as \texttt{AIC++} can guide the co-design of future heterogeneous inference platforms. From \cref{fig:deepseekv32_strict_128gpu}, we observe that under stringent SLOs for TTFT and TPOT on DeepSeek-V3.2, AFD is able to achieve ~4k tokens/s system throughput while the non-AFD deployment is infeasible to run. Through exhaustive design-space exploration to analyze the best deployment strategy on a cluster of 128 B200 GPUs (\cref{sec:eval}), we see that AFD is not the best option when system throughput is the main target compared to the pure P/D disaggregation and chunked prefill. But our analysis that dynamically searches the attention-to-FFN ratios finds that it will always achieve the best latency and user interactivity with the AFD-specific microbatch overlapping technique that best utilizes the compute and communication resources. Finally, we present a case study highlighting AFD's memory-segmentation benefit: by placing most model weights on FFN GPUs, AFD leaves more memory on attention GPUs for KV cache, enabling higher throughput under the same memory constraint.

Our key contributions are:

\begin{itemize}
\item[\textbf{C1:}] \textbf{Multi-dimensional DSE:} We jointly optimize LLM serving across token-level parallelism, P/D disaggregation, and AFD, and translate workload characteristics into concrete attention-to-FFN GPU ratios.
\item[\textbf{C2:}] \textbf{System modeling for disaggregated architectures:} We build \texttt{AIC++}, which models compute and data-transfer costs for scaled-out disaggregated inference systems.
\item[\textbf{C3:}] \textbf{Optimal AFD operator placement:} We demonstrate that careful placement of attention and FFN operators is critical for effective AFD deployment.
\end{itemize}

\section{Background and Motivation}

\subsection{Design-Space Exploration for Optimal Disaggregated Inference}
As model sizes continue to grow and new architectures emerge, the need for inference scheduling optimization has intensified, giving rise to an increasingly large design space. Finding the optimal scheduling strategy for large-scale disaggregated inference is critical to meet the stringent service-level latency and throughput requirements.

\textbf{The complex multi-dimensional search space:}
LLM inference scheduling spans a rich design space that combines multiple parallelism strategies with phase‑ and operator‑level disaggregation. Prior work~\cite{agrawal2024vidurlargescalesimulationframework, bambhaniya2026mistcodesignframeworkheterogeneous} has explored parallelization across data and model dimensions—data and sequence parallelism partition inputs, tensor and pipeline parallelism split model computation, and mixture‑of‑experts models introduce expert parallelism to scale execution across nodes (\cref{fig:dse}). In large‑scale disaggregated deployments, these strategies can be composed across input, model, and expert dimensions to improve performance. Beyond parallelism, recent systems advocate prefill–decode disaggregation~\cite{splitwisesim2024,zhong2024distserve} to accommodate heterogeneous execution characteristics, while more recent efforts propose fine‑grained operator‑level disaggregation that separates attention and FFN layers~\cite{zhu2025megascaleinfer} to better match their distinct compute and memory demands. Despite this growing set of techniques, how these dimensions interact—and when each disaggregation or parallelization strategy is most effective—remains poorly understood. In this work, we systematically study cross‑dimensional interactions in the inference design space to identify regimes where different scheduling strategies deliver the greatest benefit.

\begin{figure}[t]
    \centering
    \includegraphics[width=\linewidth]{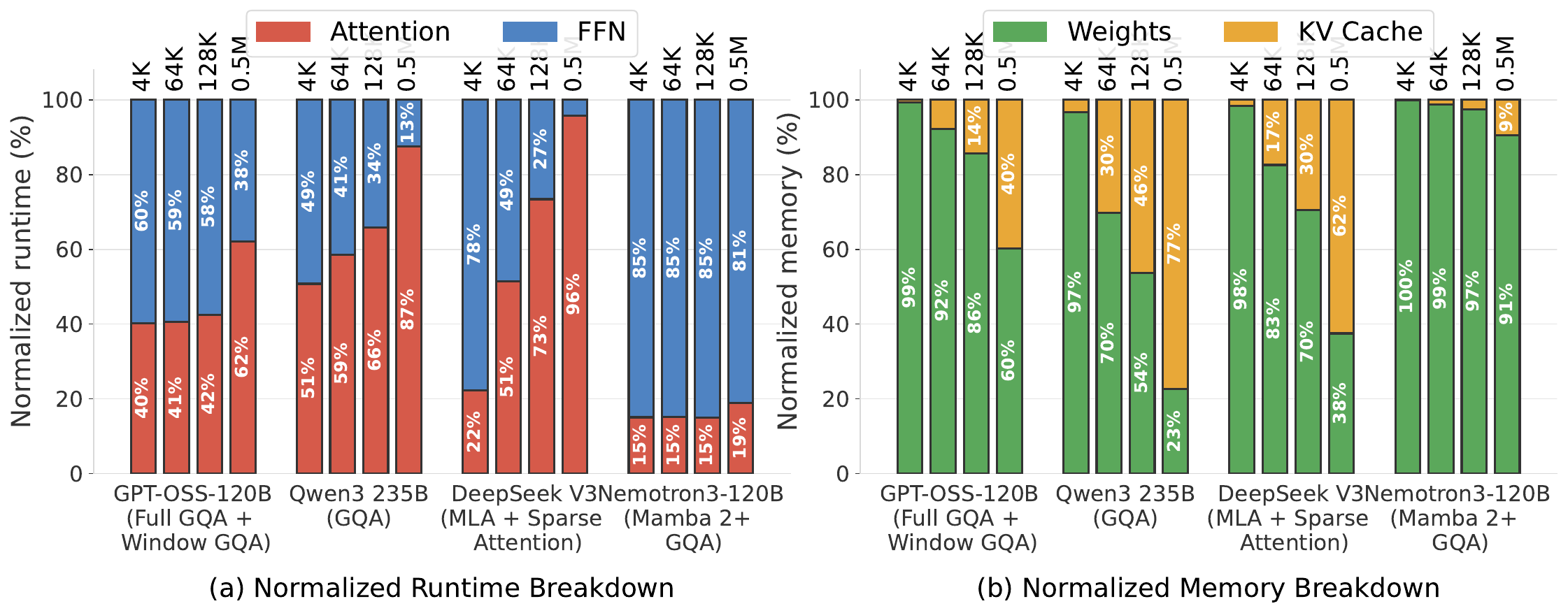}
    \vspace{-1.5em}
    \caption{(a) Runtime breakdown and (b) memory component breakdown for different model architectures running prompts with different context lengths. 
    }
    \vspace{-1.5em}
    \label{fig:model_runtime_breakdown}
\end{figure}

\textbf{Impact of model architecture on scheduling: }
The effectiveness of attention–FFN disaggregation (AFD) is highly sensitive to the underlying attention mechanism. As shown in \cref{fig:model_runtime_breakdown}(a), models with dense GQA attention (e.g., \texttt{GPT‑OSS}, \texttt{Qwen3}) are attention‑dominated at large context lengths, whereas sparse‑attention models such as \texttt{DeepSeek‑V3.2} are FFN‑dominated for short contexts and increasingly attention‑bound as context grows. In contrast, state‑space models like Mamba‑2 shift runtime decisively toward FFNs due to their linear‑time attention behavior, even at long contexts. Moreover, the attention‑to‑FFN runtime ratio increases with context length across all architectures as KV‑cache footprint grows (\cref{fig:model_runtime_breakdown}(b)). These trends highlight that attention and FFN costs are strongly workload‑ and architecture‑dependent, necessitating workload‑aware scheduling. In this work, we address this challenge through a design‑space exploration framework that explicitly captures attention characteristics and memory requirements to provision compute and communication resources accordingly. 

\textbf{Impact of system load on scheduling:}
System load, in the form of the number and size of the requests, significantly influences the scheduling strategy - especially in phase (P/D) and operator (AFD) disaggregation.
This stems from the increase in data-transfer with the increase in the number of tokens served. 
Prior application profiling reveals that the size of the input/output sequence length (ISL/OSL), the prefix length and the number of requests are dependent on the type of application.
For instance, retrieval-augmented generation (RAG) workloads - such as legal advisory chatbots - typically exhibit large context length and small ISLs.  
These workloads favor aggregated deployments, as frequent cross‑node data transfers in disaggregated systems incur high communication overheads. 
In contrast, workloads with long ISLs, common in code‑completion and code‑generation tasks, benefit more from asynchronous function decomposition (AFD), which introduces an additional dimension to the scheduling design space.
In this work, we characterize different workloads with different ISL, OSL and context length to find the Pareto-optimal scheduling that balances system throughput and user interactivity under SLO constraints (detailed in \autoref{sec:eval}). 

\subsection{Finding the optimal scheduling with disaggregated architecture}
The emergence of disaggregated architectures with heterogeneous compute units within a node—such as NVIDIA Groq‑3 LPX, Rubin CPX~\cite{Rubinmic2}, and Intel SambaNova—has made fine‑grained AFD an effective scheduling strategy despite the increased communication volume. High‑bandwidth scale‑up interconnects in these heterogeneous clusters enable memory‑bound attention operators to execute on memory‑rich devices, while compute‑intensive FFN blocks benefit from accelerators optimized for high arithmetic throughput.

\textbf{System modeling requirements of disaggregated architecture:}
Exploring a large design space by evaluating hundreds of candidate configurations is prohibitively expensive for large‑scale disaggregated systems as observed by~\cite{galvatron, alpa}.
Hence, accurate system modeling of compute and communication infrastructure is necessary for evaluation. 
AIConfigurator~\cite{xu2026aic} provides a compute modeling framework to estimate the compute and memory bandwidth of a wide range of modern GPUs.
However, we additionally need accurate estimation of the data-transfer cost for fine-grained operator-level AFD.
To address this, we augment AIConfigurator with the AstraSim~\cite{astrasim} network simulator to build \texttt{AIC++},
a disaggregated modeling framework that captures the compute-communication effects, maximizing their overlap for fine-grained AFD. 
Moreover, we evaluate the impact of AFD in heterogeneous scale-up \texttt{AIC++} deployments.

\textbf{Precise placement of attention and FFN operators:}
Fine‑grained data transfers introduced by AFD can lead to significant interconnect congestion if attention and FFN operators are placed arbitrarily across a disaggregated infrastructure. Consequently, effective scheduling of fine‑grained operator disaggregation requires jointly reasoning about compute affinity and data‑movement costs—a challenge explicitly addressed in our work. In particular, optimal operator placement must account for compute–communication overlap by co‑locating operators with high data‑exchange intensity on nearby or tightly coupled compute nodes. Going beyond prior approaches that primarily model heterogeneous architectures, our work jointly optimizes operator placement by explicitly considering interconnect congestion, identifying optimal mappings of attention and FFN operators that maximize overall system efficiency.
\section{\texttt{AIC++} Framework Overview}
In this section, we present \texttt{AIC++}, a framework for exploring the design space of AFD‑enabled MoE serving at scale. \texttt{AIC++} combines AIConfigurator~\cite{xu2026aic} for accelerator‑level performance modeling with ASTRA‑Sim~\cite{astrasim} for high‑fidelity network simulation, enabling evaluation across attention architectures and application domains. While AIConfigurator accurately models modern LLM kernels, it does not capture AFD‑specific communication paths; we extend it by partitioning execution into attention and MoE‑FFN phases and binding each phase to an independent GPU backend, faithfully modeling runtime and memory behavior on disaggregated devices.

\subsection{AFD design for MoE architectures}
As shown in~\cref{fig:Batch_overlap}, the transition between the Attention and MoE--FFN phases is mediated by the MoE-Dispatch and MoE-Combine communication operators. In non-AFD deployments, these operators involve matched source and destination counts, as communication is confined to GPU ranks participating in expert parallelism (EP). Under AFD, attention and MoE--FFN phases execute on physically disaggregated GPUs, resulting in asymmetric fan-out or fan-in communication patterns depending on the scheduling strategy. For example, while an MoE model deployed on 8 GPUs with EP=8 exchanges tokens among all 8 GPUs, an AFD configuration with 2 attention GPUs and 6 FFN GPUs induces a fan-out pattern, requiring tokens generated by the attention GPUs to be distributed to a larger set of FFN GPUs.

\texttt{AIC++} models these interactions using AstraSim to simulate scale-up and
scale-out communication at packet granularity. By coupling AstraSim with
AIConfigurator, \texttt{AIC++} enforces communication-dependent execution, capturing
contention, GPU utilization, and data-transfer efficiency under dynamic workloads.

Concretely, each transformer layer incurs two cross-AFD transfers. In the all-pairs mode used for the main results, the AFD worker establishes pairwise communicators between every attention rank and every FFN rank, so tokens generated by a single attention rank may be routed to any FFN rank hosting the selected experts.
\textbf{A2F/A2E (MoE-Dispatch)} transfers post-attention hidden states, token ids, and per-expert routing metadata from attention ranks
to FFN ranks hosting the selected experts, where each FFN rank filters tokens according to its hosted experts. Due to the top-$k$ routing and token dispatch, A2F exhibits a fan-out, making FFN-side ingress congestion dominant. \textbf{F2A/E2A (MoE-Combine)} aggregates expert outputs and
returns a single reduced hidden state per token to the originating attention rank,
forming a fan-in transfer in which attention-side ingress becomes the bottleneck. \texttt{AIC++} expands both transfers into a full bipartite
traffic matrix and feeds them into AstraSim’s tiered, congestion-aware network model,
capturing contention when A2F egress and F2A ingress overlap on full-duplex
interconnects. This part of the implementation with its communication pattern is based on our prototype implementation (refer to \autoref{sec:vllm_impl}).

\subsection{Batch Overlap (BO) in AFD}

\begin{figure}[ht]
    \centering
    \vspace{-1em}
    \includegraphics[width=\linewidth]
    {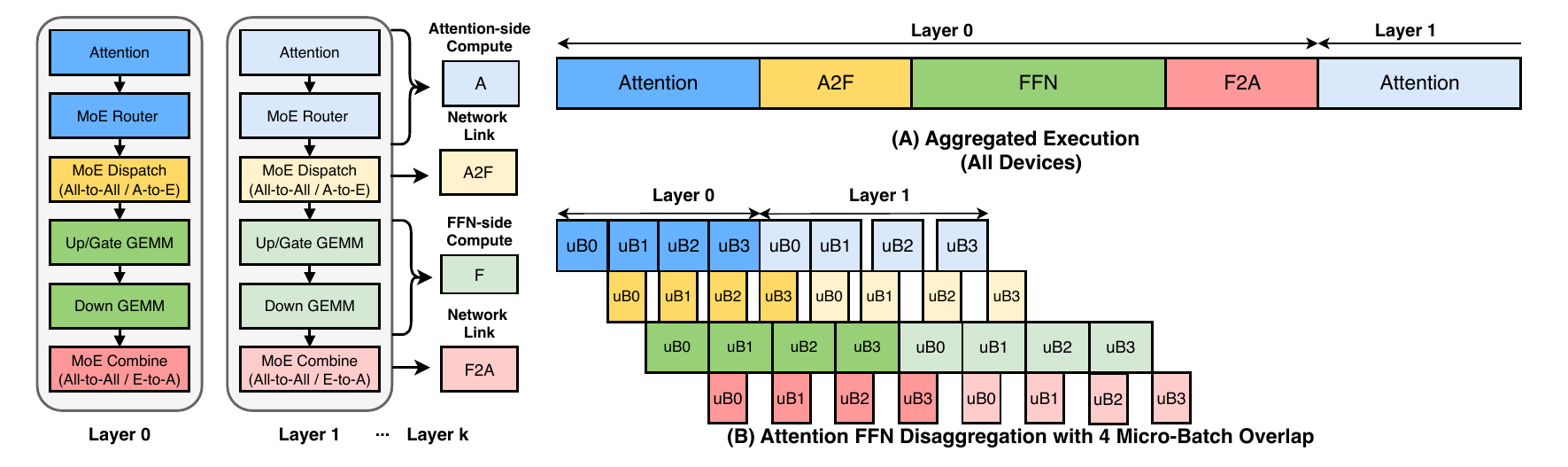}
    \vspace{-2em}
    \caption{Mapping AFD to 4 different stages in the model execution}
    \label{fig:Batch_overlap}
    \vspace{-2em}
\end{figure}

As shown in~\cref{fig:Batch_overlap} (left), AFD decomposes execution into four stages mapped to separate compute or communication resources: attention computation on the attention GPUs, dispatch of post-attention hidden states, token ids, and per-expert routing metadata to experts, MoE-FFN computation on the FFN GPUs, and aggregation/transfer of expert outputs back to the attention side for the next layer.

The return communication may share the same channel on half-duplex links, or proceed concurrently over a dedicated channel on full-duplex networks, enabling the pipelined execution shown in~\cref{fig:Batch_overlap}(B). Since modern datacenter GPU deployments commonly provide full-duplex interconnects such as NVLink and InfiniBand, AFD can exploit compute-communication overlap across GPUs and the network fabric. Accordingly, in the following discussion, we assume an AFD implementation with four-stage micro-batch overlap enabled. In contrast, under aggregated execution, all devices jointly execute all stages as a single group, as shown in~\cref{fig:Batch_overlap}(A).

To model batch overlap behavior in \texttt{AIC++}, we partition the per-step token budget
\(T_{\text{budget}} = \textit{batch\_size} \times \textit{ISL}\) into \(M\) microbatches.
The number of microbatches \(M\) is chosen to match the effective pipeline depth—three
for half-duplex links and four for full-duplex links. For each microbatch,
\texttt{AIC++} queries AIConfigurator’s empirically measured GPU-cluster cost database
to obtain the attention and FFN execution costs for a microbatch size of
\(T_{\text{budget}}/M\), thereby preserving the nonlinear scaling behavior of small
GEMMs and collective operations.

Let \(s_i\) denote the measured per-microbatch execution cost of pipeline stage \(i\),
aggregated across all \(L\) transformer layers, and let
\(s_{\max} = \max_i s_i\) be the bottleneck stage. Under steady-state cross-layer
pipelining, the bottleneck stage processes all \(M\) microbatches back-to-back,
whereas each non-bottleneck stage incurs only a one-time pipeline fill and drain
overhead of \(s_i/L\). This cost is amortized over the full execution rather than
charged per layer, consistent with the cross-layer scheduling strategies used by
MegaScale-Infer and Step-Fun. The resulting end-to-end pipelined latency is shown in~\cref{eq:bo-pipe}.
\begin{equation}
\label{eq:bo-pipe}
t_{\text{pipe}} = M \cdot s_{\max} + \sum_{i : s_i \neq s_{\max}} \frac{s_i}{L}.
\end{equation}
The first term captures the steady-state throughput dictated by the bottleneck
resource, while the second term accounts for pipeline fill and drain bubbles
introduced by non-bottleneck stages. We apply this formulation to both prefill and
decode phases, enabling \texttt{AIC++} to accurately reason about computation and
communication costs under batch-overlapped execution.

\subsection{Location-aware GPU placement}

Because AstraSim labels each GPU with its physical position—node, scale-up domain, and link tier—and resolves congestion at packet granularity, \texttt{AIC++} additionally enables a study of optimal GPU placement under combined AFD and P/D disaggregation. The policy is frequency-driven: the most frequent intra-layer A2F/F2A traffic ($\mathcal{O}(\text{layer})$ per request) is bound to the highest-bandwidth scale-up domain (intra-node NVLink), while the less frequent inter-node KV-cache transfer ($\mathcal{O}(1)$ per request) is deferred to the scale-out domain (InfiniBand). This grouping co-locates GPUs with high communication affinity onto faster interconnects, avoiding contention on slower links. The detailed placement study—including segregated vs.\ paired P/D layouts and the resulting KV-transfer speed-up—is given in Appendix~\ref{app:gpu-placement}.

We also consider asymmetric prefill and decode configurations reflecting their distinct compute characteristics. \texttt{AIC++} enables such asymmetry through independent prefill/decode scheduling and asymmetric attention and FFN worker allocation, while the network simulator co‑locates operators to reduce interconnect congestion and improve efficiency.


\section{Evaluation}\label{sec:eval}
\subsection{Design-Space Exploration (DSE) of different serving strategies}

\subsubsection{Input workloads}
To demonstrate the applicability of AFD, we evaluate a diverse set of model architectures spanning multiple application domains, as summarized in~\cref{tab:model_app_workloads}. Specifically, we consider recent production‑deployed models including \texttt{Qwen3‑235B}, commonly used for chatbot workloads; \texttt{GPT‑OSS‑120B}, targeting medium‑scale reasoning tasks; \texttt{DeepSeek‑V3.2}, designed for reasoning‑intensive and agentic coding workflows; and \texttt{Nemotron3‑120B}, optimized for large‑context applications such as RAG serving~\cite{nemotron3}. The corresponding prefix sizes, ISL, OSL, and architectural characteristics for each workload are detailed in~\cref{tab:model_app_workloads}.

\begin{table}[ht]
\vspace{-1em}
\caption{Representative application workloads and model architectures.}
\centering
\label{tab:model_app_workloads}
\scriptsize
\setlength{\tabcolsep}{3pt}
\renewcommand{\arraystretch}{1.12}
\begin{minipage}[t]{0.42\columnwidth}
\vspace{0pt}
\centering
\textbf{Application Workload}

\vspace{0.35em}
{\renewcommand{\arraystretch}{1.35}%
\begin{tabular}{@{}lccc@{}}
\toprule
\textbf{Use Case} & \textbf{Prefix} & \textbf{ISL} & \textbf{OSL} \\
\midrule
\textbf{Chat}           & 4096   & 512 & 256 \\
\textbf{Coding}         & 2048   & 4096  & 1024  \\
\textbf{Agentic Coding} & 524k \footnotemark  & 256 & 8192  \\
\bottomrule
\end{tabular}}
\end{minipage}
\begin{minipage}[t]{0.56\columnwidth}
\vspace{0pt}
\centering
\textbf{Model Architecture}

\vspace{0.35em}
\begin{tabular}{@{}llcl@{}}
\toprule
\textbf{Model} & \textbf{Attention} & \textbf{\# Experts} & \textbf{Precision} \\
\midrule
\textbf{GPT-OSS-120B}   & Full GQA + Window GQA  & 128 & FP8 \\
\textbf{Qwen3-235B}     & GQA                    & 128 & FP8 \\
\textbf{Nemotron3-120B} & Mamba 2 + GQA          & 512 & FP8 \\
\textbf{DeepSeek V3.2}  & MLA + Sparse Attention & 256 & FP8 \\
\bottomrule
\end{tabular}
\end{minipage}
\hspace{0.002\columnwidth}
\vspace{-2em}
\end{table}

\footnotetext{Models listed in this table may not natively support a 524k
context window. We use 524k to model long-context agentic workloads and quantify
the system impact of large KV-cache residency.}

\subsubsection{Cluster-scale DSE analysis}
\begin{figure}[t]
    \centering
    \includegraphics[width=\linewidth]{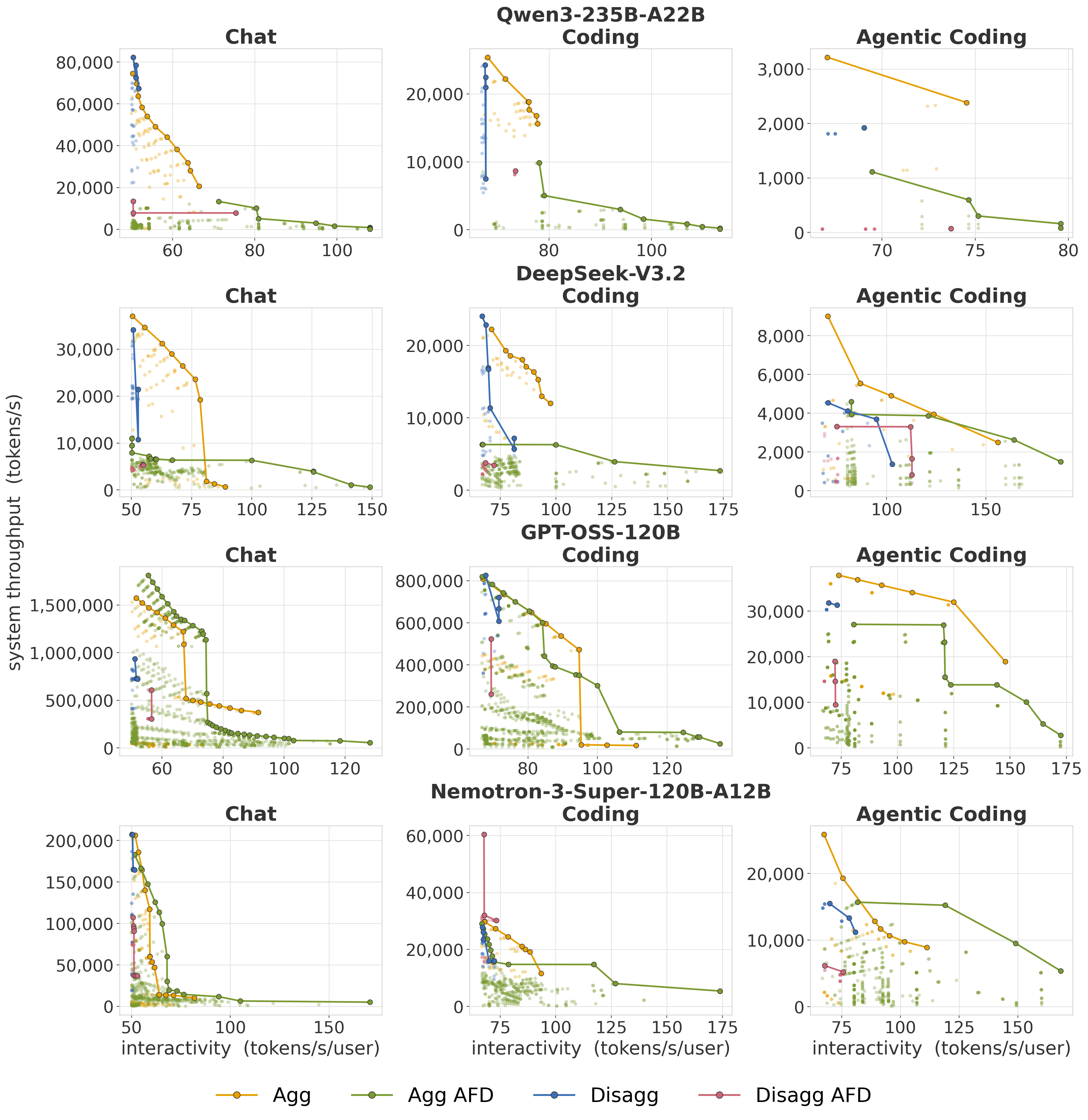}
    \vspace{-1.5em}
    \caption{Evaluation of a Cluster with 128 B200 GPUs for the representative model architectures and application workloads shown in \autoref{tab:model_app_workloads}, Y-axis denotes the system throughput (total tokens/s), X-axis denotes the interactivity (tokens/s/user)}  
    \vspace{-1.5em}
    \label{fig:chat_code_agent_pareto}
\end{figure}

\cref{fig:chat_code_agent_pareto} reports the Pareto curve (\texttt{tokens/s/user} vs system \texttt{tokens/s}) produced by \texttt{AIC++} on a 128 B200\textsubscript{SXM} cluster with TensorRT-LLM~\cite{tensorrt_llm} as the performance backend, with each panel constrained by the workload's SLO from \cref{tab:model_app_workloads}. To exhaustively probe the throughput frontier, our replica search enumerates every replica size from 2 to 128 GPUs and dynamically composes the per-replica parallelism plan across TP, DP, EP, and (for AFD) attention/FFN GPU groups. This lets the optimizer surface asymmetric layouts that only become feasible when prefill and decode have very different compute and memory profiles, including off-grid replica sizes that pack a long-context KV cache more efficiently than the obvious power-of-two choices. The cluster-scale results in this section are model-based DSE estimates that combine backend cost measurements with AstraSim communication modeling. Appendix~\ref{sec:vllm_impl} describes our vLLM-based AFD prototype, which we use to verify functional correctness of the all-pairs attention--FFN execution path and to ground the communication pattern modeled by \texttt{AIC++}.

\textbf{No single strategy dominates the throughput frontier.} Aggregated serving with chunked prefill, deployed as 16 single-node 8-GPU replicas that hold the expert FFNs through Expert Parallelism, wins most panels by amortizing the chunked-prefill bubble across the replica fleet and ingesting many disjoint token batches in parallel. Disaggregation takes the rest, but only after the wider search exposes asymmetric replica shapes: a single off-grid replica with many small 2-GPU prefill workers feeding a few large 8-GPU decode workers (Qwen3-235B-A22B chat, DeepSeek-V3.2 coding), or many tiny xPyD shards under \emph{disagg+AFD} that double aggregated's throughput on Nemotron-3-Super coding. AFD on its own wins the throughput frontier on a single panel, GPT-OSS-120B chat, where the cheap sliding-window GQA layers let four 32-GPU replicas with a near-symmetric 16A+16F split outpace 16-replica agg. The wider TP-16 enumeration also unlocks a feasible \emph{disagg-standard} layout for DeepSeek-V3.2 agentic coding, where a narrower TP,$\le$,8 search returned no SLO-feasible configuration for the 524k-prefix workload.

\textbf{On the latency axis, AFD wins every panel}, with the optimal attention/FFN split tracking each model's intrinsic attention/mixer cost relative to its FFN cost. \textbf{DeepSeek-V3.2}, whose MLA combined with sparse (DSA) attention shrinks both per-token attention compute and the KV-cache footprint, collapses the attention shard to its minimum on long-context workloads, dedicating almost the entire cluster to FFN (2A+126F on agentic, 16A+112F on chat). The 2A+126F layout is initially counter-intuitive given the 524k-token KV demand, but consistent with rate-matching: MLA compresses the latent KV cache so aggressively that the entire prefix fits in two GPUs' HBM, and DSA keeps per-token attention cheap enough that two attention GPUs running large concurrent decode batches sustain the rate emitted by a 126-GPU FFN pool; growing the attention slice would idle it. \textbf{Qwen3-235B-A22B}'s dense GQA demands a wider attention slice (8A+120F on agentic, growing as TTFT tightens), reflecting that its attention path keeps up with FFN only when allocated more compute. \textbf{GPT-OSS-120B}'s alternating sliding-window GQA sits in the cheap-attention regime, tolerating a small attention shard on long prefixes, yet on chat the same architecture lets AFD win on throughput with a near-symmetric 16A+16F split, a behavior absent in the other three. \textbf{Nemotron-3-Super-120B} inverts the pattern: its Mamba2+GQA hybrid mixer must propagate a recurrent state across the 524k-token prefix, and the throughput-optimal AFD layout flips attention-heavy (96A+32F); under tighter TTFT budgets it returns to an FFN-heavy split once the bottleneck shifts from state propagation back to FFN matmul.

The Attention-to-FFN Split ratio follows a rate-matching scheme: AFD allocates attention GPUs only as far as needed to match the FFN's output rate, parameterized by per-token attention cost and KV/state memory. Cheap, low-memory attention (MLA+DSA, sliding-window GQA) sustains a large FFN pool from a small attention slice, pushing 90+\% of the cluster to FFN; heavy attention (like dense full-context GQA's growing KV cache) or lower cost MoE FFN cannot, and the ratio shifts to scale attention up. 

\textbf{Key Takeaway 1 - System Throughput: } \textit{Aggregated wins by data-parallel concurrency across fully-replicated workers; disagg/AFD wins when independent sizing of prefill/decode or attention/FFN better rate-matches their compute than a uniform replica.}

\textbf{Key Takeaway 2 - User Interactivity: }\textit{AFD wins user interactivity universally by sizing the attention-to-FFN ratio per workload and model to achieve low latencies through micro-batch overlapping}

\subsubsection{Long Context Case Study}

Building on the Cluster-DSE analysis, and motivated by the growing context-window demands of agentic workloads, we further study a long-context use case with \(\mathrm{ISL}=500\mathrm{k}\) + \(\mathrm{OSL}=10\mathrm{k}\) and another case with \(\mathrm{Prefix}=1\mathrm{M}\) + \(\mathrm{ISL}=4\mathrm{k}\) + \(\mathrm{OSL}=500\). We evaluate Qwen3-235B on B200 in HYBRID database mode with AFD and a 4-stage microbatch pipeline (\(M=4\)). For each scenario and GPU budget, we search over valid model-parallel layouts and report the feasible configuration that achieves the highest total token throughput under the TTFT/TPOT constraints. Aggregated serving is searched in 8-GPU worker-size increments to match the B200 scale-up unit, while disaggregated serving separately searches prefill and decode workers and then rate-matches the two phases.

\begin{figure}[t]
    \centering
    \includegraphics[width=\columnwidth]{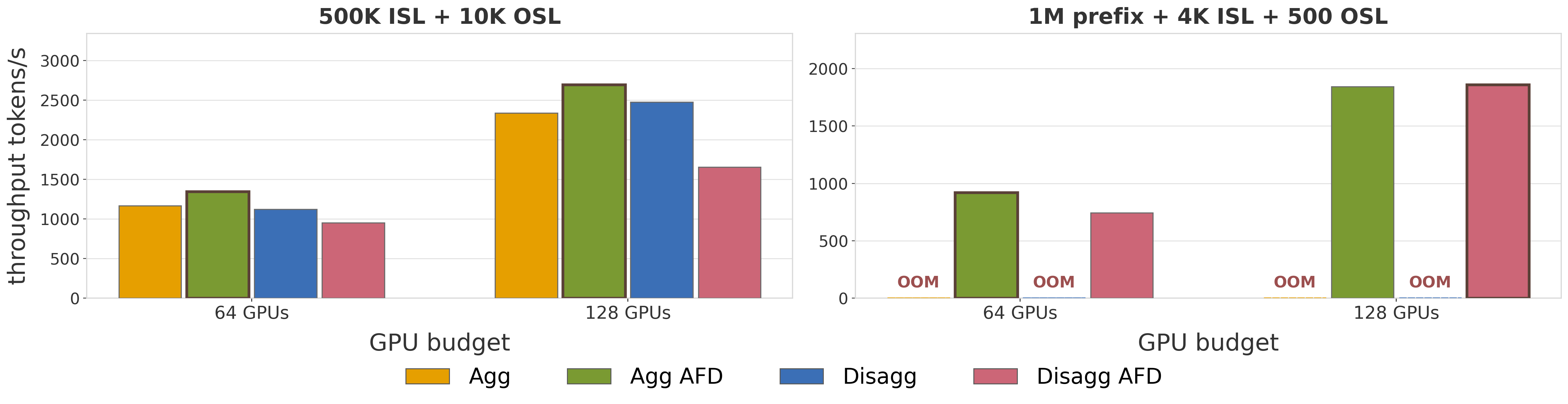}
    \vspace{-8pt}
    \caption{Throughput frontier for the long-context Qwen3-235B workload on B200 GPUs.}
    \label{fig:longctx_qwen3}
    \vspace{-2em}
\end{figure}

\autoref{fig:longctx_qwen3} compares two long-context regimes. The left subplot shows a large-prefill workload with \(\mathrm{ISL}=500\mathrm{k}\) and \(\mathrm{OSL}=10\mathrm{k}\), which is strongly prefill-dominated. In this setting, \texttt{Agg+AFD M4} achieves the highest throughput at both GPU budgets, reaching \(1346.5\) tok/s at 64 GPUs and \(2693.0\) tok/s at 128 GPUs. The winning layouts are 2 replicas of \([28A+4F]\) at 64 GPUs and 4 replicas of \([28A+4F]\) at 128 GPUs, corresponding to an attention-to-FFN split of \(7{:}1\). This allocation is favorable because long-prefill requests are attention- and KV-dominated: assigning more GPUs to the attention side improves long-context processing, while the \(M=4\) microbatch pipeline overlaps the AFD stages and further increases throughput.

The right subplot shows a large-prefix regime with \(\mathrm{prefix}=1\mathrm{M}\), \(\mathrm{new\ input}=4\mathrm{k}\), and \(\mathrm{OSL}=500\). In this regime, non-AFD modes are infeasible because the per-GPU memory requirement,
\(M_{\mathrm{shared}} = W{+}A{+}K{+}N{+}O \approx 298\,\mathrm{GiB}\),
including weights, activations, KV cache, NCCL buffers, and runtime overhead, exceeds the 180-GiB B200 memory capacity. AFD reduces the memory pressure by splitting weights and activations across the attention and FFN GPU groups, lowering the effective per-GPU requirement to
\(M_{\mathrm{AFD}} = \max(M_{\mathrm{attn}}, M_{\mathrm{ffn}}) \approx 165\,\mathrm{GiB}\),
which fits within the device memory budget. Under this setting, \texttt{Agg+AFD M4} achieves the highest throughput at 64 GPUs (\(921.5\) tok/s), while \texttt{Disagg+AFD M4} narrowly wins at 128 GPUs (\(1858.9\) vs.\ \(1843.0\) tok/s).

\section{Discussion and Conclusion}\label{sec:conclusion}

We study fine-grained attention--FFN disaggregation (AFD) using \texttt{AIC++} to navigate the joint space of parallelism, P/D disaggregation, and AFD. AFD lowers latency by sizing the attention and FFN pools independently, with the optimal ratio set by the workload applications and the model's attention mechanism. The trade-off is concurrency: each AFD replica consumes more GPUs, so aggregated layouts retain the throughput crown on most panels, and AFD's value is strongest at latency-sensitive points and on long-context regimes where aggregated mode exceeds the per-GPU memory cap. The emerging trend toward heterogeneous in-node accelerators directly motivates this primitive; extending \texttt{AIC++} beyond B200 to unreleased accelerators and NPUs is left to future work.

\bibliographystyle{plainnat}
\bibliography{Sections/refs}
\clearpage
\section{Appendix}

\subsection{vLLM Implementation for AFD}\label{sec:vllm_impl}

We prototype AFD on top of vLLM v0.16.0rc2~\cite{kwon2023vllm} to verify
functional correctness of the full bipartite attention--FFN execution path and
to align the implementation structure with the communication pattern modeled in
\texttt{AIC++}. The prototype is used for correctness and implementation
guidance, while the cluster-scale results use measured backend costs and
AstraSim communication modeling. Our starting point is an open-source vLLM AFD implementation~\cite{vllm_pr29772}, which supports
MoE models (DeepSeek V2/V3, Step3) as an opt-in runtime mode with an NCCL
P2P transport and a 1:1 attention$\leftrightarrow$FFN pairing that delegated expert
routing to an intra-FFN all-gather plus reduce-scatter. This baseline only
supports \emph{symmetric} configurations ($M = N$), which excludes design
points worth exploring at scale, e.g.\ 1A4F, 2A4F, and the $x$A1F family. To
exercise arbitrary attention/FFN ratios, we re-architect the AFD implementation.
Specifically, we introduce:

\paragraph{(i) $M\times N$ bipartite pairing.}
We replace the 1:1 zip with a full bipartite product over $(M, N)$: every attention
rank is paired with every FFN rank via its own NCCL pair-group, for a total of
$MN$ pairs. Pair-group bootstrap uses a per-pair Gloo rendezvous on unique TCP
ports so that the $MN$ NCCL communicators can be constructed deadlock-free
without contaminating the global process group's counter state. The topology
is valid for arbitrary $M, N \ge 1$ and subsumes the original symmetric case.

\paragraph{(ii) Attention-side routing with bipartite transport.}
We relocate the MoE router (gate projection, top-$k$ selection) and the
shared experts from the FFN side to the attention side. Each attention
rank computes routing decisions locally and transfers the post-attention
hidden state, token ids, and per-expert routing metadata to each paired
FFN partner over the corresponding NCCL pair-group. Each FFN partner
runs only its local expert shard and returns a \emph{partial} tensor;
the attention side sums the partials element-wise across FFN partners
and adds the locally-computed shared-expert output. The MoE path
therefore involves \emph{no FFN$\leftrightarrow$FFN collective}: all
inter-worker communication is point-to-point on the $M\times N$
bipartite graph---the same send/receive pattern adopted by production
AFD-targeted communication libraries such as StepMesh~\cite{stepmesh}.
We support two transport modes on this topology. The \emph{dense} mode
ships the full post-attention payload over every link and lets
\texttt{expert\_map} zero out unmatched tokens on the FFN side, which
is sufficient when the per-link payload is small. When the dense
payload becomes large, an opt-in \emph{sparse pre-routing} mode has each
attention rank pre-mask on GPU and send only the tokens whose top-$k$
experts hit a local expert of the receiving FFN, with a single batched
non-blocking host transfer per layer reading all per-partner counts in
one shot. Both modes produce bit-identical FFN outputs given identical
routing.

For modeling at scale, \texttt{AIC++} reconciles this implementation by
inheriting the same all-pairs bipartite topology---every attention rank
flows to every FFN rank, so the traffic matrix is fully populated. With
$M$ senders and $N$ receivers, the per-link load is balanced when
$M = N$, while $M > N$ congests the receiver side (FFN ingress) and
$M < N$ congests the sender side (attention egress). On each
attention-to-FFN edge we charge an expected post-filter payload under
uniform top-$k$ routing across FFN ranks, plus a small routing-metadata
overhead; the F2A return is modeled symmetrically. AstraSim then
resolves topology, contention, and link-level effects over this
expected-payload bipartite traffic matrix.

\paragraph{(iii) Expert-kernel bypass.}
In the original vLLM entry point, \lstinline|FusedMoE.quant_method.apply|
dispatches into \lstinline|FusedMoEModularKernel.forward|, whose
\lstinline|prepare_finalize| hooks themselves invoke EP all-gather and
reduce-scatter---the very collectives that attention-side routing is designed to
avoid. We therefore introduce a new entry point
\lstinline|FusedMoE.forward_pre_routed| that calls the raw
\lstinline|fused_experts| Triton kernel directly with \lstinline|expert_map|
filtering, yielding a zero-collective MoE-layer implementation on the FFN
side.


\subsection{GPU-aware operator placement with network enhanced simulation}
\label{app:gpu-placement}

\begin{figure}[ht]
    \centering
    \includegraphics[width=8.5cm]{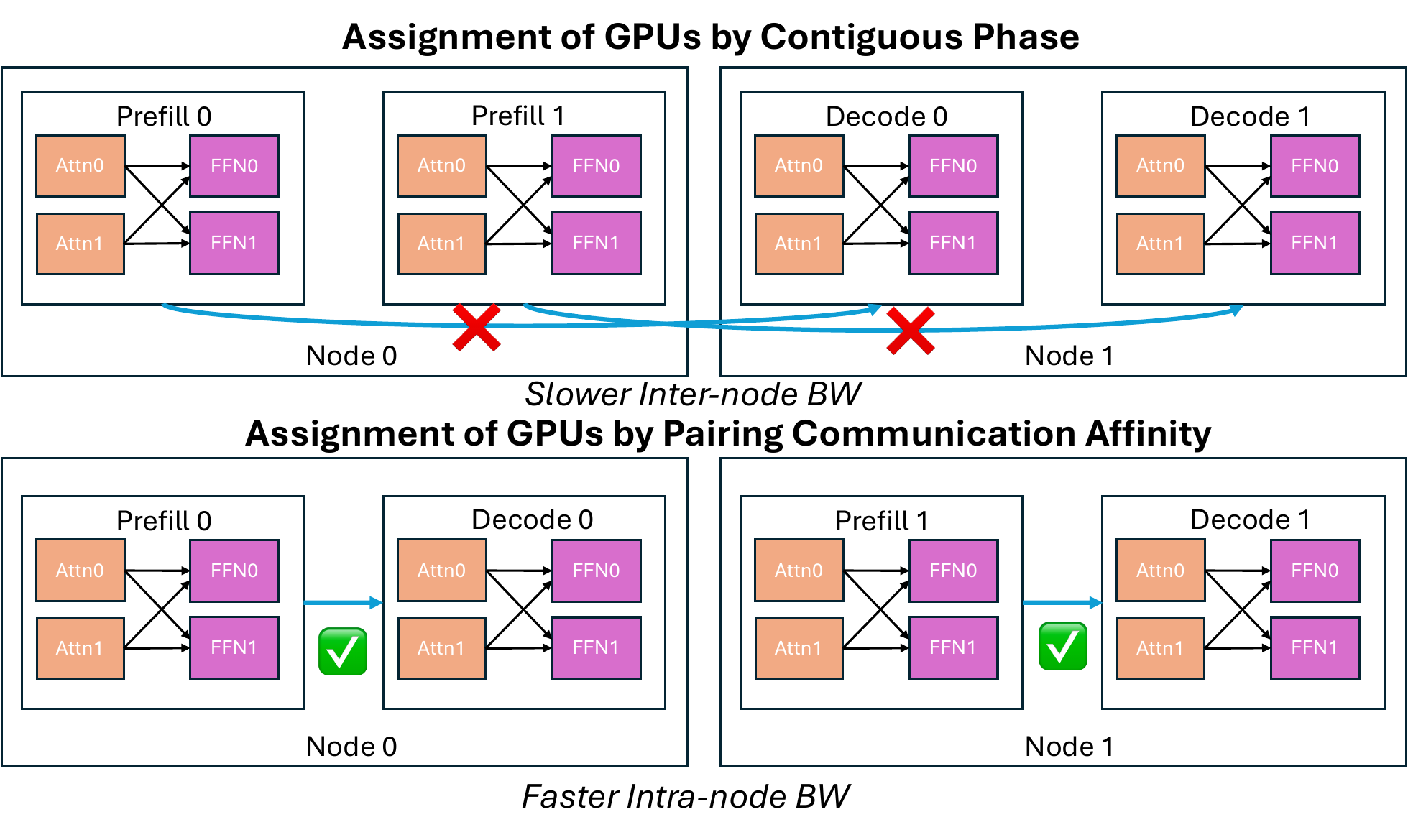}
    \vspace{-8pt}
    \caption{GPU Placement}
    \label{fig:GPU_Placement}
\end{figure}

Combining AFD with P/D disaggregation in disaggregated serving requires multiple co-existing communication patterns at different granularity.
\textbf{(1) Per-layer MoE dispatch/combine} ($\mathcal{O}(\text{layer})$ per request) requires data transfer between the attention and the MoE-FFN GPUs for each layer; and \textbf{(2) KV-cache transfer} occurs once per request, when the prefill worker ships the materialized KV cache to the decode GPU.
Since the per-layer AFD transfer is invoked more frequently and scales with the number of layers, we assign the attention and the FFN operators in the scale-up domain (intra-node NVLink) to minimize the communication cost, and defer the less-frequent prefill-to-decode KV transfer to the scale-out domain (inter-node InfiniBand) as illustrated in~\cref{fig:GPU_Placement}.

\begin{table}[ht]
\centering
\small
\begin{tabular}{l c c c}
\toprule
\textbf{Worker} & $\text{G}_{\text{attn}}$ & $\text{G}_{\text{ffn}}$ & \textbf{Node} \\
\midrule
\multicolumn{4}{l}{\textit{(a) Segregated by phase} — all prefill first, then decode} \\[2pt]
$\text{Prefill}_0$ & $\{0,1\}$    & $\{2,3\}$    & 0 \\
$\text{Prefill}_1$ & $\{4,5\}$    & $\{6,7\}$    & 0 \\
$\text{Decode}_0$  & $\{8,9\}$    & $\{10,11\}$  & 1 \\
$\text{Decode}_1$  & $\{12,13\}$  & $\{14,15\}$  & 1 \\
\midrule
\multicolumn{4}{l}{\textit{(b) Paired P\&D per node} — co-locate each $(P_i, D_i)$} \\[2pt]
$\text{Prefill}_0$ & $\{0,1\}$    & $\{2,3\}$    & 0 \\
$\text{Decode}_0$  & $\{4,5\}$    & $\{6,7\}$    & 0 \\
$\text{Prefill}_1$ & $\{8,9\}$    & $\{10,11\}$  & 1 \\
$\text{Decode}_1$  & $\{12,13\}$  & $\{14,15\}$  & 1 \\
\bottomrule
\end{tabular}
\caption{GPU placement for 2A2F with 2P2D on two 8-GPU nodes.
In~(a), KV transfers cross the node boundary;
in~(b), each prefill--decode pair shares a node.}
\label{tab:gpu-placement}
\end{table}

\begin{figure}[ht]
    \centering
    \includegraphics[width=8.5cm]{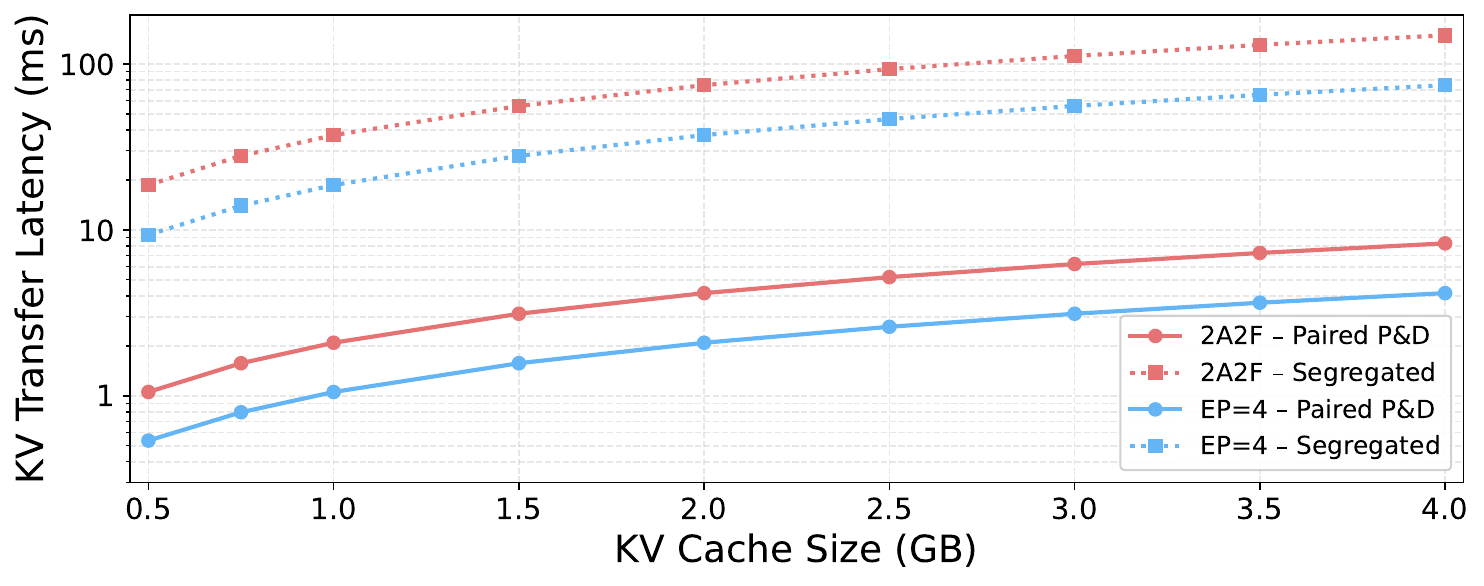}
    \vspace{-8pt}
    \caption{KV transfer latency under 2P2D: segregated by phase (dotted)
vs.\ paired P\&D per node (solid), for 2A2F (red) and EP\,=\,4 (blue).
Latency scales linearly with KV size (from 500MB to 4GB).}
    \label{fig:kv-placement-study}
\end{figure}

We study a \textbf{2A2F} (equiv. EP=4) configuration ($N_{\text{attn}}=2$,
$N_{\text{ffn}}=2$ per worker) in a \textbf{2P2D} layout across two
8-GPU nodes. Table~\ref{tab:gpu-placement} shows both placement
strategies and the corresponding network link used for KV-cache
transfers. After forming the communicator group, we model the
rail-optimized switch network architecture~\cite{nvidia_gtc24_railoptimized}.
Under the segregated placement, KV transfers cross the node boundary
over InfiniBand at $BW_{\text{IB}}=25$\,GB/s; under the paired
placement, they remain on NVLink at $BW_{\text{NVLink}}=450$\,GB/s,
yielding an $18\times$ speed-up as shown in
Figure~\ref{fig:kv-placement-study}. While our example targets the system with scale-up size of 8, this placement analysis
generalizes naturally to any multi-tiered network hierarchy---the same
priority-ordered assignment applies whenever the topology exposes
tiers with distinct bandwidths (e.g., NVL72 with NVLink, NVSwitch,
and InfiniBand), always binding the most frequent communication
pattern to the fastest available link.

\end{document}